\documentclass[11pt]{article}

\pdfoutput=1

\usepackage{NLPAICS2024}

\usepackage{times}
\usepackage{latexsym}
\usepackage{dirtytalk}
\usepackage{enumitem}
\usepackage{booktabs}
\usepackage{adjustbox}

\usepackage[T1]{fontenc}

\usepackage[utf8]{inputenc}

\usepackage{microtype}

\usepackage{inconsolata}

\title{Explainability of machine learning approaches in forensic linguistics:\\a case study in geolinguistic authorship profiling}

\author{Dana Roemling \\
  University of Birmingham \\
  University of Helsinki \\
  \texttt{\small d.roemling@bham.ac.uk} \\\And
  Yves Scherrer \\
  University of Oslo \\
  University of Helsinki \\
  \texttt{\small yves.scherrer@ifi.uio.no}  \\\And
  Aleksandra Miletić \\
  University of Helsinki \\
  University Sorbonne Nouvelle\\
  \texttt{\small aleksandra.miletic@helsinki.fi} \\}

\begin{document}
\maketitle
\begin{abstract}
Forensic authorship profiling uses linguistic markers to infer characteristics about an author of a text. This task is paralleled in dialect classification, where a prediction is made about the linguistic variety of a text based on the text itself. While there have been significant advances in recent years in variety classification, forensic linguistics rarely relies on these approaches due to their lack of transparency, among other reasons. In this paper we therefore explore the explainability of machine learning approaches considering the forensic context. We focus on variety classification as a means of geolinguistic profiling of unknown texts based on social media data from the German-speaking area. For this, we 
identify the lexical items that are the most impactful for the variety classification. We find that the extracted lexical features are indeed representative of their respective varieties and note that the trained models also rely on place names for classifications.
\end{abstract}

\section{Introduction}

Forensic authorship analysis is a key area of research within forensic linguistics that seeks to gain information about the author(s) of a text. Generally, there are two central domains of research within authorship analysis: \emph{comparative authorship analysis} uses various methodologies to compare questioned and known documents to attribute authorship, while \emph{sociolinguistic} or \emph{authorship profiling} relies on the analysis of questioned texts alone to infer characteristics of an author \citep{Grant_2022, Roemling_Grieve_2024}. Both areas of authorship analysis can be approached qualitatively and/or quantitatively if the amount of available data permits it. For example, quantitative work in authorship profiling has researched inferring age or gender \citep{Nini_2018} or native language influence \citep{Kredens_Perkins_Grant_2019} from questioned documents. 

Nevertheless, forensic authorship profiling is often carried out in a manual or qualitative way, relying on the expertise of the forensic linguist. This is evident in famous examples like the \emph{Unabomber} analysis \citep{Leonard_Ford_Christensen_2017} or the \emph{devil strip} ransom note \citep{Shuy_2001}. In both cases, law enforcement was interested in the regional origin of the authors. This background can be inferred through analyzing the regional linguistic variation, i.e., the use of regional dialect, in the questioned documents. This is referred to as \emph{regional} or \emph{geolinguistic profiling} \citep{Roemling_Grieve_2024} and is a task parallel to inferring the regional variety of a document as is done in language identification \citep{Jauhiainen_Lui_Zampieri_Baldwin_Lindén_2019}. 

Even though research in forensic linguistics works more and more with statistical and computational approaches \citep[e.g.,][]{Bevendorff_Chinea-Ríos_Franco-Salvador_Heini_Körner_Kredens_Mayerl_Pęzik_Potthast_Rangel_etal._2023, Ishihara_Kulkarni_Carne_Ehrhardt_Nini_2024, Nini_Halvani_Graner_Gherardi_Ishihara_2024}, authorship profiling often remains a manual task. This is at times credited to the black-box approaches in current NLP research, meaning that the lack of explainability precludes these approaches from being used in legal settings \citep[see][]{Nini_2023}.

\section{Related work}

The interest in explainability of machine learning (and in particular, neural) approaches is not only a relevant research area for forensic linguistics. Previous work, including on language identification, has focused on understanding how classifiers come to their predictions. Research started by creating an interpretable model around the actual classification approach to explain the predictions of the classifier \citep{Ribeiro_Singh_Guestrin_2016}. \citet{Li_Monroe_Jurafsky_2017} employed representation erasure to propose a methodology for interpretability research while \citet{Jacovi_SarShalom_Goldberg_2018} explored how filters in CNNs can be understood in text classification research, finding that filters do in fact learn different classes of ngrams. Furthermore, \citet{Ehsan_Tambwekar_Chan_Harrison_Riedl_2019} showed that by training on human explanation data, models can learn to translate their inner states into understandable explanations. \citet{Belinkov_Gehrmann_Pavlick_2020} thus summarised the main sub-fields of interest in interpretability research as focusing on \say{probing classifiers, behavioral studies and test suites, and interactive visualizations}.

\citet{Xie_Ahia_Tsvetkov_Anastasopoulos_2024} were the first to research explainability in a dialect classification context. In order to analyze the classifier, they extracted lexical features that were highly relevant to the classification, aiming to use this knowledge for dialect research and not only as a means to explore machine learning approaches. They relied on lexical items, owing to the complex nature of handling preprocessing like tokenisation or POS-tagging in low(er)-resourced varieties. They indicated that refined approaches would be beneficial. However, previous research on regional variation using social media data has shown that approaches using lexical features provided excellent results \citep[e.g.,][]{Doyle_2014, Huang_Guo_Kasakoff_Grieve_2016, Eisenstein_2017, Grieve_Nini_Guo_2018, Grieve_Montgomery_Nini_Murakami_Guo_2019}. 

\citet{Xie_Ahia_Tsvetkov_Anastasopoulos_2024} proposed two different approaches, one intrinsic and one post-hoc, to extract features relevant to the dialect classification. In the intrinsic approach, the authors added a local interpretability layer to the dialect classifier which was trained together with the model and output the relevance of a feature for the classification. For the post-hoc approach, \citeauthor{Xie_Ahia_Tsvetkov_Anastasopoulos_2024} used a leave-one-out (LOO) method, where the change in prediction probability if a feature was left out was interpreted as a \say{relevance score} of that particular feature. 

In a forensic authorship profiling setting, an approach like this appears beneficial as it reaches high accuracies in language identification, while similarly providing explanations by extracting the features that influenced the classification. While the original study focused on improving research methods in dialectology, we apply the method to evaluate its usefulness in a forensic context. Additionally, the approach has the advantage of eliminating or at least minimising researcher bias as the models learn the relevant features themselves, whereas it is the forensic linguist's expertise that culls the features in a qualitative analysis \citep[see][]{Grant_Grieve_2022}\footnote{Note, however, that the training data itself may introduce bias into the model \citep[see][]{Blodgett_Barocas_DauméIII_Wallach_2020}.}. While the approach does not fully explain the inner workings of the model, experts can use the extracted features to a) verify that the model indeed reached a sound decision, for example by evaluating the features against previous dialectological findings, and b) use the explanations to introduce the method to law enforcement or jurisprudence. Even if the classifiers themselves do not meet court admissibility standards \citep{Coulthard_2013, Hammel_2022}, extracted features can be used for authorship work to contribute to a more objective analysis. 

\section{Data}
We work with a corpus of German social media data from the platform Jodel. The corpus was collected by \citet{Hovy_Purschke_2018, Purschke_Hovy_2019}. It has also been used in VarDial classification tasks \citep{Gaman_Hovy_Ionescu_Jauhiainen_Jauhiainen_Lindén_Ljubešić_Partanen_Purschke_Scherrer_etal._2020, Chakravarthi_Gaman_Ionescu_Jauhiainen_Jauhiainen_Lindén_Ljubešić_Partanen_Priyadharshini_Purschke_2021}. Jodel is structurally similar to Twitter/X, however it only allows anonymous posts. Users of Jodel can interact with other users in a 10-15 km radius around their own location, so all posts are geolocated. The corpus contains posts from Austria, Germany and Switzerland. While most of the data is written in standard German, it shows clear regional patterns. Especially in Switzerland, Austria and, in parts, Bavaria writing is considerably further from standard German \citep{Purschke_Hovy_2019}. Posts from Romandy contain substantial amounts of French. This data differs from the corpus used in the original study \citep[see][]{Xie_Ahia_Tsvetkov_Anastasopoulos_2024} in terms of register and genre.

\begin{figure} 
\includegraphics[width=\columnwidth]{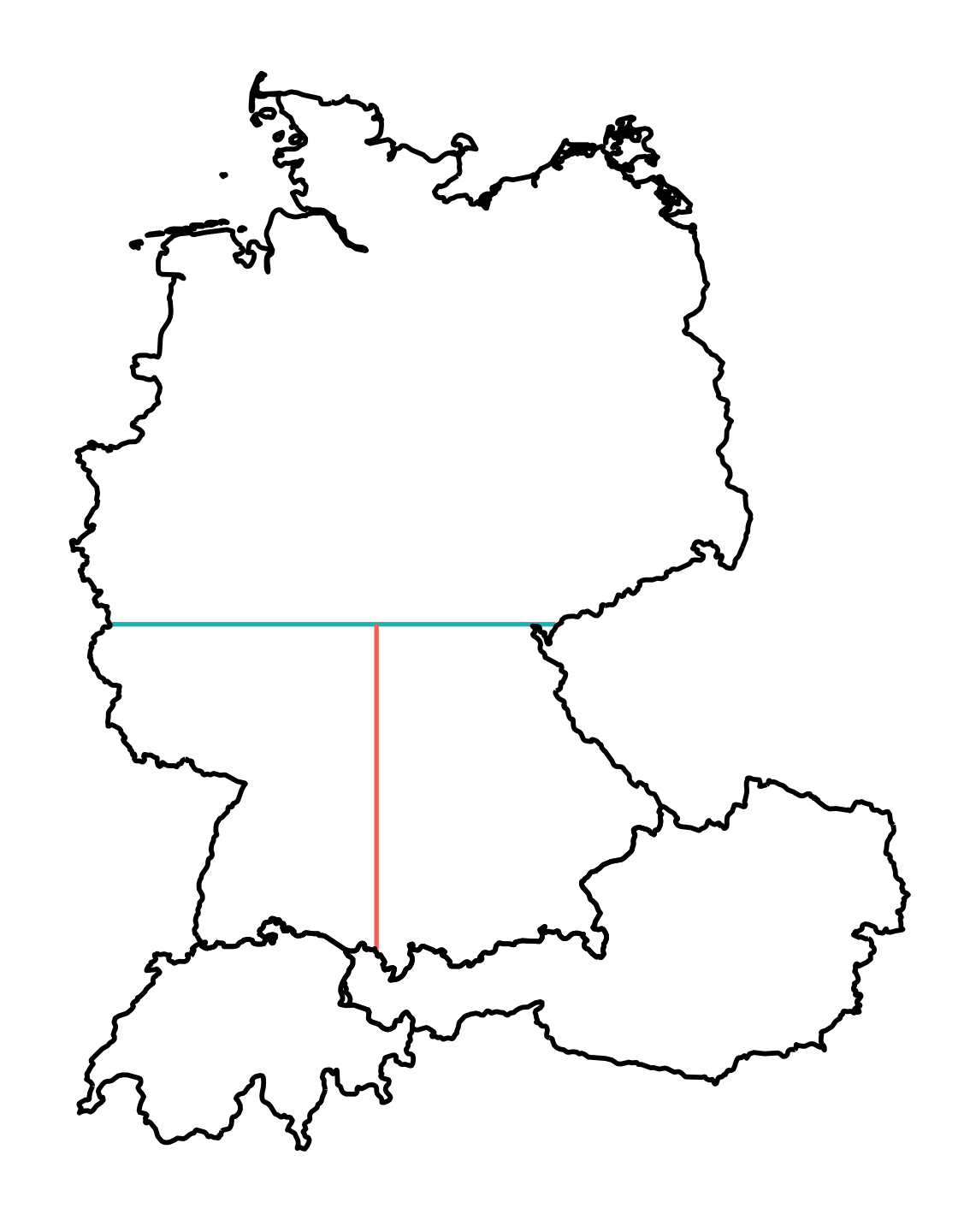}
\caption{Operationalization of dialect regions}
\label{fig:operationalization}
\end{figure}

The corpus consists of approximately 240 million tokens from about 8500 locations, however only 388 locations have a token count of over 10k. For our classification experiments, we mapped these locations onto wider dialect regions following three settings with 3, 4 and 5 classes respectively. The 3-class distinction is based on national borders, so the classes reflect Austria, Germany and Switzerland. In the 4-class setting Germany is additionally divided into two parts, north and south (at latitude 50.33° N) and for the 5-class setting the southern region of Germany is further split into east and west (at longitude 9.97° E) (see Figure \ref{fig:operationalization}). These divisions were operationalized based on knowledge from traditional dialectology \citep{Wiesinger_1983, König_2004}.

We randomly sampled 200k posts per class for training, and 20k posts per class for development and testing, respectively. On average, a post contains 11.5 tokens. Besides some simple whitespace normalization (i.e., removing line breaks and tabs inside a post), we did not perform any preprocessing.

\section{Methods and experiments}

The basis for our analysis is the post-hoc LOO approach proposed by \citet{Xie_Ahia_Tsvetkov_Anastasopoulos_2024}. While we replicate most of the methodology, we mark any changes from the original in our explanations below. Crucially, and in contrast to the original dialectal research interest, we apply the approach with the goal of evaluating its usefulness in a forensic setting. Additionally, we work with data that is different in terms of register and genre and thus adds to our overall understanding of the approach.

\subsection{Dialect classifiers}

The approach described by \citet{Xie_Ahia_Tsvetkov_Anastasopoulos_2024} starts by fine-tuning a BERT-based language model on the dialect classification task. Following \citet{Xie_Ahia_Tsvetkov_Anastasopoulos_2024}, we rely on the multilingual model \texttt{xlm-roberta-base}\footnote{\url{https://huggingface.co/FacebookAI/xlm-roberta-base}} \citep{conneau-etal-2020-unsupervised} but we also experiment with a base model specifically trained on German data, \texttt{dbmdz/bert-base-german-cased}\footnote{\url{https://huggingface.co/dbmdz/bert-base-german-cased}}.

Considering two base models and three settings (3, 4 and 5 classes), our experiments yield six dialect classifiers. The training is done with the \texttt{simpletransformers} library\footnote{\url{https://simpletransformers.ai/}}. Each model is trained for 10 epochs with a maximum sequence length of 256 (subword tokens) and a batch size of 64 samples. We use default values for all other parameters.

\begin{table}
\adjustbox{max width=\linewidth}{%
\begin{tabular}{lrrr}
\toprule
Classes & Random & XLM-RoBERTa & German BERT \\
\midrule
3 & 33\% & 75.31\% & 74.91\% \\
4 & 25\% & 58.57\% & 58.44\% \\
5 & 20\% & 47.74\% & 47.64\% \\
\bottomrule
\end{tabular}}
\caption{Classification accuracies on the development sets.}
\label{tab:class_accuracy}
\end{table}

Table~\ref{tab:class_accuracy} shows the classification accuracies of these models on the development sets. It can be seen that all models outperform the random baseline by a large margin. The difference between the two base models is marginal and we thus focus on the German base model.

\subsection{Leave-one-word-out classification}

The LOO method used by \citet{Xie_Ahia_Tsvetkov_Anastasopoulos_2024} processes each sentence of the test set independently to detect the words that contribute most to the classification. It consists of the following steps:

\begin{enumerate}[nolistsep]
\item Select an instance $x$ of the test set, run it through the dialect classifier, and record the predicted class $\hat{y}$ as well as the prediction score $\ell$. If the prediction is incorrect ($\hat{y} \neq y$), skip this instance and move to the next one.\footnote{This corresponds to the \textit{isCorrect} constraint of \citet{Xie_Ahia_Tsvetkov_Anastasopoulos_2024}.}
\item Select one word of the instance and remove it from the instance (let $x_i$ denote the instance $x$ from which word $i$ is removed), run it through the dialect classifier, and record the prediction score $\ell_i$.\footnote{\citet{Xie_Ahia_Tsvetkov_Anastasopoulos_2024} select the word to be removed by position, with the consequence that if a word occurs several times in the same sentence, only one of its occurrences will be removed at a time. They then only consider the occurrence that produced the biggest difference. We simplify this part by iterating over the set of unique words and removing all occurrences of the selected word at the same time.}
\item Measure the impact of the removed word on the classification performance ($\Delta_i$) by subtracting the score of the incomplete instance from the score of the complete instance: $\Delta_i = \ell - \ell_i$. We call $\Delta_i$ the \textit{impact score} of word $i$.
\item Repeat steps 2 and 3 for each word of the sentence.
\item Select the 5 words with the highest impact score.\footnote{\citet{Xie_Ahia_Tsvetkov_Anastasopoulos_2024} omit this step in the description of their work, but it is present in their code.}
\end{enumerate}

The steps described above produce \textit{explanations} at instance level, i.e., the most impactful words of each instance. In a forensic case setting with limited data, this could already be leveraged by using the impactful features in a qualitative analysis or simply evaluating and explaining the prediction made by the classifier. Consequently, this can be an interesting result in itself, but for the analysis reported here, we aggregate the explanations across all instances of the test set for evaluation. The resulting list is processed in the following way:

\begin{enumerate}[nolistsep]
\item Words that were selected as explanations for more than one class are eliminated from further consideration.\footnote{This corresponds to the \textit{isUnique} constraint of \citet{Xie_Ahia_Tsvetkov_Anastasopoulos_2024}.}
\item Words that figure as explanations for only one instance are eliminated from further consideration.
\item For each remaining word, we compute the average impact score on the basis of the individual impact scores.\footnote{\citet{Xie_Ahia_Tsvetkov_Anastasopoulos_2024} use TF-IDF to rank the words. We find that the simpler approach of averaging the scores is sufficient for our purposes.}
\item For each class, we select the 100 words with the highest average impact scores for the analysis.
\end{enumerate}

\section{Results}
Following the method described above, we produce a list of 100 words with the highest impact scores per class in all settings. A manual inspection of the lists quickly shows a prevalence of place names and related items like \textit{Zürich} or \textit{Österreicher} (G `Austrian')\footnote{Throughout the remainder of the paper, the examples from Germany are marked with \emph{G}, those from Austria with \emph{AG}, and those from Switzerland with \emph{SWG}. Examples from Romandy are marked \emph{FR} for French.}. Therefore, as a first step, we count the amount of these words among the top 100 and find that, on average, 14\% of words are local references. In terms of classification, these results are expected \citep[see, e.g.,][]{Nasar_Jaffry_Malik_2022} and it is apparent how these words are indicative of location given what they denote. Although it does not take a forensic linguist to understand the connection of these items to location, it is noteworthy that the classification models pick up on these words and that they match the region they are impactful for. 

Given that the results are similar for classes based on country borders across the three settings, we focus our analysis on the 5-class setting. As a reminder, in this setting, Austria and Switzerland form individual classes, whereas Germany is split into three regions (see Figure~\ref{fig:operationalization}). Generally, the results show that a large proportion of extracted words are regionalized and some are prototypical dialect items. 

For Switzerland, we find both Swiss German and French lexical items to be the most impactful, such as \textit{Kei} (SWG `not a/no'), \textit{Isch} (SWG `is') and \textit{bim} (SWG `at the'), or \textit{pourquoi} (FR `why'), \textit{Avec} (FR `with') and \textit{raison} (FR `reason'). Whereas for Austria, the data shows that some of the most impactful lexical items are \textit{Oasch} (AG `ass'), \textit{Gspusis} (AG `affairs'), and \textit{Matura} (AG `high-school diploma'). These three items are examples of a textualization of regional pronunciation, a regional item and a standard Austrian variant, respectively. Also, items such as \textit{Jus} (AG `law (studies)'), which are relevant to the Jodel demographic of mostly students under the age of 27, are extracted. This indicates that an automated classification between the three countries, for instance to clarify jurisdiction, seems reasonable.

For the three German classes we find that a large proportion of top words appear textualized in standard German as opposed to more colloquial spellings including abbreviations and ellipses, which we may expect. Examples of this include \textit{Dankeschön} (G `thank you') or \textit{Vorname} (G `first name'). Considering the division within Germany, we find that several forms of the verb \textit{gucken} (G `to look/watch') are impactful for the northern class, which is a variant we know to be regionalized and appearing in varieties in central and northern Germany \citep[][p. 235]{König_2004}. For the south-east class the data shows items like \textit{Ritter} (G `knight'), \textit{\#traudel} (female first name) and local beer types. For the south-west we find that items identified as relevant by \citet{Purschke_Hovy_2019}, are also extracted by the LOO model, like \textit{Möppes} (G `breasts' or `female user') and \textit{Lörres} (G `penis' or `male user'), although the authors argue that these forms are more Jodel- than region-specific. These items are also impactful for the Germany-class in the 3-class setting. 

\section{Conclusion}
In this paper, we employed \citet{Xie_Ahia_Tsvetkov_Anastasopoulos_2024}'s approach as a means of geolinguistic profiling to understand how a method like this could be applied in a forensic context given its explainability. We have found that the dialect classifiers outperform the random baseline by a large margin in all settings, but that accuracy decreases for settings with more closely-related classes. While we recognize that in forensic contexts the focus needs to be on false predictions and hard-to-classify cases, this paper considers the explainability of the approach. To this end we have found that the LOO model extracts meaningful regional features reflecting the variety used in the corresponding area. On average, 14\% of extracted features are place names or similar items. While an analysis for a place name like \textit{Wien} does not need a forensic linguist, extracting features that are not based on a linguist's expertise is a valuable contribution of this approach even if it is not directly used for automated classification.

\section*{Limitations}
For this paper we have worked with the geolocation of the Jodel posts as the gold label for the dialect regions used as classes in the classification task. However, there is noise in this data as people move and use varieties from different regions in the same place. Further analysis of the incorrect classifications may allow us to identify these instances. 

For further work it may be beneficial to remove non-German varieties before training. Additionally, given the high percentage of place names and related lexical items, preprocessing to remove named entities \citep[see, e.g.,][]{Darji_Mitrović_Granitzer_2023} may help focus the extraction on dialectal lexical items. 

\section*{Acknowledgements}
Dana Roemling was supported by the UKRI ESRC Midlands Graduate School Doctoral Training Partnership ES/P000711/1. Yves Scherrer and Aleksandra Miletić were supported by the Academy of Finland through project No.~342859 ``CorCoDial -- Corpus-based computational dialectology''.

We thank Dirk Hovy and Christoph Purschke for sharing their data with us. We also wish to acknowledge CSC – IT Center for Science, Finland, for computational resources.

\bibliography{anthology,custom}
\bibliographystyle{acl_natbib}

\end{document}